\pdfoutput=1

\documentclass[11pt]{article}

\usepackage[]{ACL2023}

\usepackage{times}
\usepackage{latexsym}

\usepackage[T1]{fontenc}

\usepackage[utf8]{inputenc}

\usepackage{microtype}

\usepackage{inconsolata}

\usepackage{xcolor}
\usepackage{algorithm}
\usepackage{algorithmic}
\usepackage{booktabs}
\usepackage{tabularx}
\usepackage{graphicx}
\usepackage{amsfonts}
\usepackage{amsmath}
\usepackage{amssymb}

\title{Fourier Transformer: Fast Long Range Modeling\\by Removing Sequence Redundancy with FFT Operator}

\author{
Ziwei He$^\diamondsuit$, 
Meng Yang$^\dagger$, 
Minwei Feng$^\dagger$, 
Jingcheng Yin$^\dagger$, 
\\
\textbf{Xinbing Wang$^\diamondsuit$}, 
\textbf{Jingwen Leng$^\diamondsuit$} 
\and \textbf{Zhouhan Lin$^\diamondsuit$\Thanks{Zhouhan Lin is the corresponding author.}}
\\
$^\diamondsuit$Shanghai Jiao Tong University
\quad $^\dagger$ Netease BizEase
\\
\{ziwei.he, xwang8, leng-jw\}@sjtu.edu.cn
\quad $^\ast$lin.zhouhan@gmail.com
}

\begin{document}
\maketitle
\begin{abstract}
The transformer model is known to be computationally demanding, and prohibitively costly for long sequences, as the self-attention module uses a quadratic time and space complexity with respect to sequence length. Many researchers have focused on designing new forms of self-attention or introducing new parameters to overcome this limitation, however a large portion of them prohibits the model to inherit weights from large pretrained models.
In this work, the transformer's inefficiency has been taken care of from another perspective. We propose Fourier Transformer, a simple yet effective approach by progressively removing redundancies in hidden sequence using the ready-made Fast Fourier Transform (FFT) operator to perform Discrete Cosine Transformation (DCT). Fourier Transformer is able to significantly reduce computational costs while retain the ability to inherit from various large pretrained models.
Experiments show that our model achieves state-of-the-art performances among all transformer-based models on the long-range modeling benchmark LRA with significant improvement in both speed and space. For generative seq-to-seq tasks including CNN/DailyMail and ELI5, by inheriting the BART weights our model outperforms the standard BART and other efficient models. \footnote{Our code is publicly available at \url{https://github.com/LUMIA-Group/FourierTransformer}}
\end{abstract}

\section{Introduction}
Transformers \citep{vaswani2017attention}, especially when equipped with large-scale pre-training \citep{devlin2018bert, lewis2019bart, raffel2020exploring} have become the core architecture in most tasks in natural language processing (NLP), including both encoder-only tasks such as sentence classification, sequence tagging \citep{liu2019roberta}, and encoder-decoder tasks such as text summarization and question answering \citep{lewis2019bart}. However, due to the quadratic complexity of its self-attention module \citep{lin2017structured}, applying these models on long sequences can be prohibitively costly. As a result, great efforts have been put into developing various efficient Transformer variants \citep{tay2020efficient}, as well as establishing standardized test-beds for long sequences such as the Long Range Arena (LRA) \citep{tay2020long}.

Most efficient Transformers devise special attention variants to lower its complexity \citep{tay2020efficient}. Some of them achieve this by projecting components in self-attention into its lower-rank approximations \citep[\emph{inter alia}]{wang2020linformer, zhu2021long, winata2020lightweight}, or rely on kernelization to implicitly compute the attention matrix \citep[\emph{inter alia}]{katharopoulos2020transformers, choromanski2020rethinking, peng2021random, choromanski2020masked}.

Due to the introduction of projection matrices or extra parameters, these models are not able to inherit pre-trained model parameters. However, since pre-trained large language models (LLMs) have fundamentally influenced the NLP community, deviating model architecture from LLMs requires pre-training from scratch on the designed model, which is prohibitively resource-demanding for most practitioners. 

Other approaches target at computing part of the attention matrix, by following some predefined patterns \citep[\emph{inter alia}]{child2019generating, qiu2020blockwise, ho2019axial}. Some of them allow the pattern to be learnable \citep[\emph{inter alia}]{sukhbaatar2019adaptive, roy2021efficient}. Most of the patterns require customized CUDA kernels or special operators to achieve the claimed speedup \citep{wu2019pay, child2019generating, beltagy2020longformer}, which casts extra challenge in deploying these models on edge devices or special hardware such as TPUs. Moreover, some of the approaches involve considerable additional computation steps, which in practice could counterweight the time and memory complexity they reduce, especially for short and medium-length sequences \citep{kitaev2020reformer, roy2021efficient}. 

\begin{figure*}[ht]
\centering
\includegraphics[width=\textwidth]{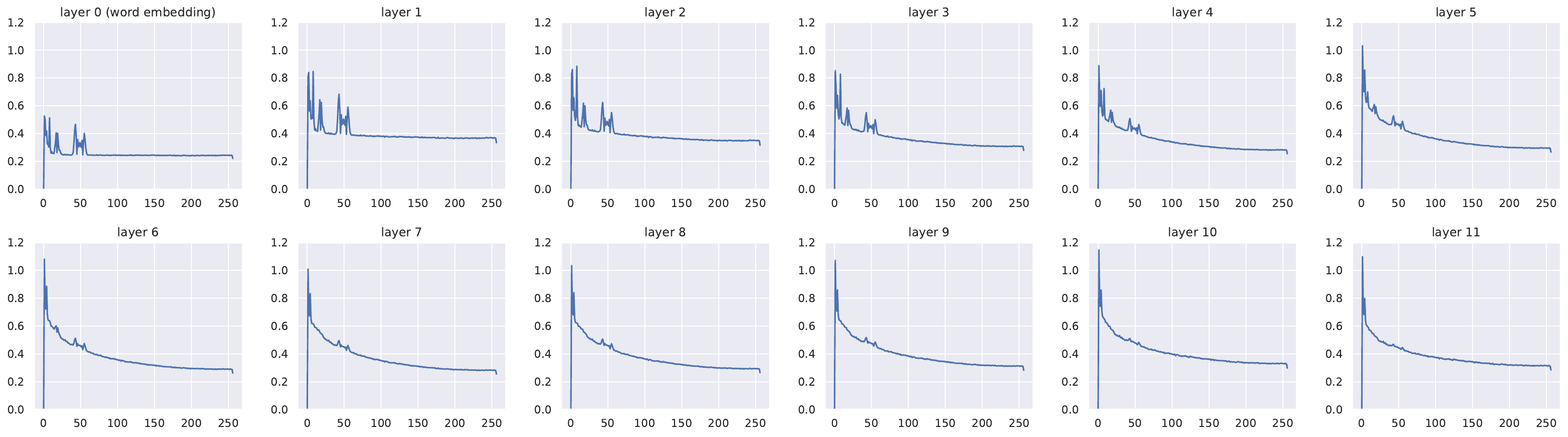} 
\caption{The power spectrum of input hidden states from different layers in the pretrained RoBERTa \citep{liu2019roberta} model. The horizontal axes stand for frequency bins, starting from low frequency components on the left. The vertical axes are the corresponding amplitudes. Amplitudes are averaged over all hidden dimensions and over the entire validation set of Wiki-103 \citep{merity2016pointer}. Since the inputs are real numbers, the positive and negative frequency components are pairwise conjugate. Thus we only plot the amplitude of the positive half of the frequencies. }
\label{fig: roberta_hiddens}
\end{figure*}

One core factor behind various approaches is the existence of redundancy in attention matrices and hidden states. For example, \citet{wang2020linformer} provides spectrum analysis on the self-attention matrix, indicating that the attention matrix learns to be low-rank, which allows them to learn a low-rank approximation of the attention matrix. Inspired by this line of research, in this work, we analyze the power spectrum of the hidden states in the time dimension through different layers in Fig \ref{fig: roberta_hiddens}, and show that the power spectrum increasingly concentrates on lower frequency bins as the layer gets deeper. 

In this work, we propose Fourier Transformer, which doesn't even require to learn the projection matrix in order to approximate the self-attention. Fourier Transformer leverages our observation on power spectra of hidden states, it progressively removes sequence redundancies through different layers by downsampling hidden states with the Discrete Cosine Transform (DCT), a variant of Fourier transform that generates real values. 

The DCT in our proposed Fourier Transformer can be implemented with the Fast Fourier Transform (FFT) operator. Thanks to its profound application in image compression and signal processing, it is one of the most widely available and highly optimized operators in a wide variety of frameworks and even on edge devices, providing $O(n\log n)$ complexity and up to $O(\log n)$ in parallel implementations with negligible overhead. As a result, Fourier Transformer is easily deployable on a wide range of devices, not necessary to devise special CUDA kernels. In addition, experimental results on LRA tasks show that it performs significantly faster than many other efficient Transformers, while achieving the state-of-the-art performance among Transformer-based efficient models.

On the other hand, since DCT is a linear, reversible transformation, and the self-attention is not interfered in our model, the proposed Fourier Transformer can inherit pretrained weights from large language models without hurting performance. Experimental results on CNN-DailyMail \citep{hermann2015teaching} and ELI5 \cite{eli5_lfqa} show that our model could outperform BART \cite{lewis2019bart} and other efficient Transformers by inheriting and fine-tuning on BART. Moreover, with tiny amount of further pretraining before fine-tuning, its performance could be further improved.

\section{Related Work}
\paragraph{Downsampling hidden states} There are not many work that downsample sequence length for natural language. The closest work is Funnel Transformer \citep{dai2020funnel}, which progressively reduces the \textit{query} sequence length through strided mean pooling, while keeping \textit{key} and \textit{value} sequence lengths intact. Fourier Transformer compresses the three sequences altogether and delivers more computational speedup compared with Funnel Transformer. Note that Funnel Transformer needs to re-invest the saved computations to build a larger model to achieve better performance, which disables its ability to inherit pretrained weights. 
For other work, Charformer \citep{tay2021charformer} devises a differentiable tokenization module that also relies on strided mean pooling to downsample its byte sequence. Nystr\"{o}mformer \citep{xiong2021nystromformer} approximates the attention matrix through the Nystr\"{o}m method, which effectively downsamples \textit{query} and \textit{key} sequences. Due to the extra depth-wise convolution, it is again not able to leverage pretrained models. 

In a border view, downsampling has been more favorable in computer vision. \citep{chen2020generative} aggressively downsamples the raw input to a 1D vector. Perceiver \citep{jaegle2021perceiver} adopts an asymmetric attention mechanism to distill inputs into a tight latent bottleneck. Almost all of these vision models are designed for encoder-only vision tasks rather than encoder-decoder-style NLP tasks. 

\paragraph{Fourier transform for Transformer}
There are multiple recent works that incorporate Fourier transform into Transformer. FNet \citep{lee2021fnet} takes a more radical approach by replacing the entire self-attention with 2D FFT, discarding the entire imaginary part to avoid complex numbers. Performer \citep{choromanski2020masked} introduced orthogonal random Fourier features to approximate the softmax attention. FSAT \citep{zhuang2022long} uses 1D FFT along the sequence dimension to learn the sparse structure of attention matrix. DCTFormer \citep{scribano2022dct} translates sequences into frequency domain and conducts self-attention there before projecting them back, due to the nonlinearity in the network, self-attention trained in the frequency domain significantly deviates from that in the time domain. Therefore, all the models discussed above lose the ability to inherit pretrained weights as well.

\section{Preliminaries}



\subsection{Discrete Cosine Transform}
The Discrete Cosine Transform (DCT) expresses a sequence of real numbers in terms of a sum of cosine functions with different frequencies. Since DCT only yields real values, it is a substitution for Fourier transform in the field of real numbers. It has been the core transform behind the JPEG \footnote{https://jpeg.org/jpeg/} lossy image compression format. 

Formally, for a sequence of $N$ real numbers $\{x_n\}=\{x_0, x_1, ... x_{N-1}\}$, DCT transforms it into frequency domain through\footnote{There are several slightly different variants of DCT. Here we use the most common type-II variant in this paper.}:
\begin{gather}
\label{eq: dct}
    y_{k} = \alpha_k \sum_{n=0}^{N-1} x_{n}cos\left( \frac{\pi k (2n+1)}{2N} \right)
\end{gather}
where $k \in \{0, ..., N-1\}$ and $\alpha_k$ is an coefficient related to $k$:
\begin{equation}
\alpha_k = 
    \begin{cases}
        \sqrt{ \frac{1}{N}} \qquad if \;\; k = 0, \\
        \sqrt{ \frac{2}{N}} \qquad otherwise \\
    \end{cases}
\end{equation}

The original sequence $\{x_n\}$ can be recovered with the inverse DCT (IDCT):
\begin{equation}
    \label{eq: idct}
    x_n = \sum_{k=0}^{N-1} \alpha_k y_k \cos\left( \frac{\pi k (2n+1)}{2N} \right)
\end{equation}
which we'll note as $\{x_n\} = IDCT(\{y_k\})$.

Practically, DCT can be computed by using the FFT operator. First, let $\{u_n\}$ be the shuffled $\{x_n\}$ by interleaving its values on even and odd positions. Formally, when $N$ is an odd integer, $\{u_n\}$ is given by
\begin{equation}
\{u_n\} = \left\{ x_{0}, x_{2}, ..., x_{N-1}, x_{N-2}, x_{N-4},..., x_{1} \right\}
\end{equation}
When $N$ is even, a similar shuffling applies. We then transform $\{u_n\}$ into its frequency domain through FFT:
\begin{equation}
\{v_k\} = FFT(\{u_n\})
\end{equation}
where $k \in \{0, ..., N-1\}$ and $\{v_k\}$ is a sequence of length $N$. The DCT of the original sequence $\{x_n\}$ can thus be computed from $\{v_k\}$:
\begin{equation}
\label{eq: dct_fft}
    y_{k} = cos\left( \frac{\pi k}{2N} \right)\mathfrak{Re}\left(v_k\right) - sin\left(\frac{\pi k}{2N}\right)\mathfrak{Im}\left(v_k\right)
\end{equation}
where $\mathfrak{Re}\left(\cdot\right)$ and $\mathfrak{Im}\left(\cdot\right)$ stand for the real and imaginary part, respectively.

\begin{figure*}[ht]
\centering
\includegraphics[width=\textwidth]{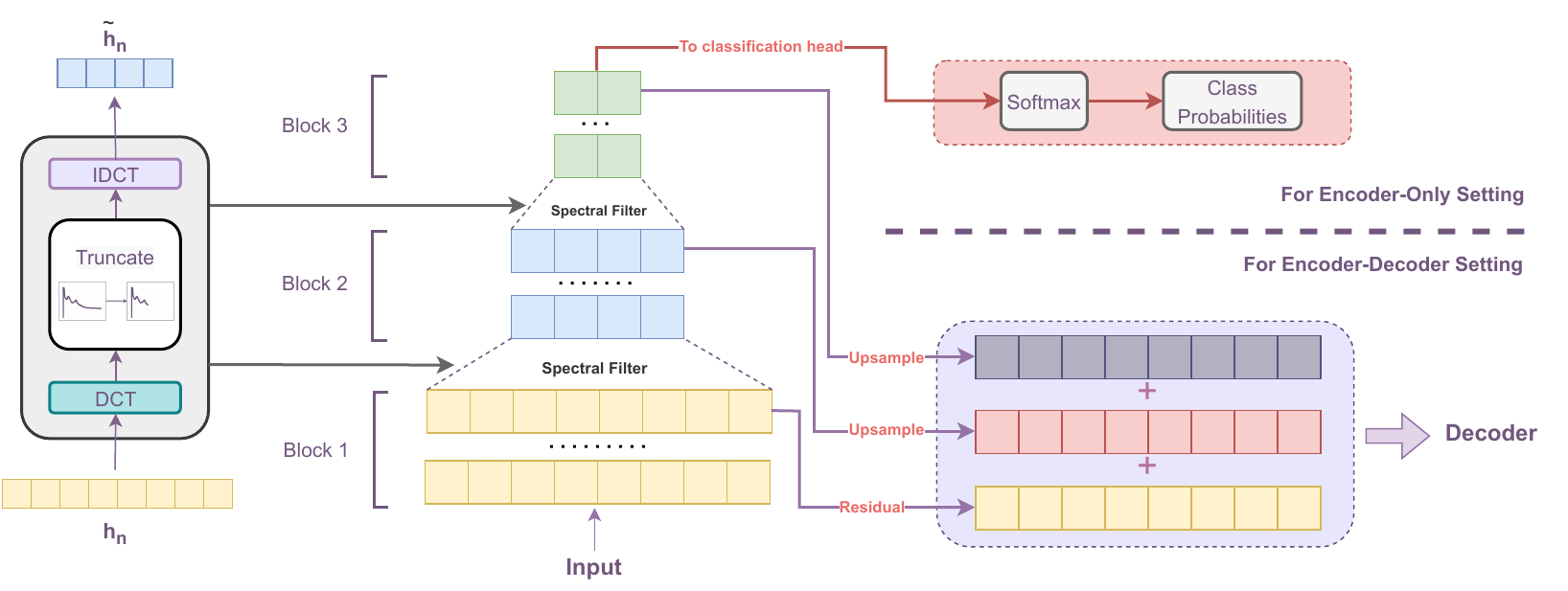} 
\caption{Overall Model Architecture}
\label{fig: model_arch}
\end{figure*}

\subsection{The Power Spectrum of Transformer Hidden States}
The power spectrum of a discrete sequence describes the distribution of signal power w.r.t. frequency components, which is the amplitudes of frequency components yielded by the Fourier transform. For a certain layer in Transformer, its hidden states can be considered as a sequence of hidden vectors, along the time dimension. To analyze the power spectrum of the layer, we conduct 1D Fourier transform independently along the time dimension for the hidden vectors, calculate the corresponding amplitudes, and avreage over all dimensions in that layer. In addition, we calculate the mean spectrum over many text sequences to eliminate example-wise noise. 

Figure \ref{fig: roberta_hiddens} shows the power spectra for different layers in the pre-trained RoBERTa-base \citep{liu2019roberta} model. The up-left subfigure shows that the power spectrum of word embeddings is relatively flat, distributing its energy almost uniformly on all frequency components with several spikes in low frequencies. As the layer gets deeper, the energy starts to concentrate toward low frequencies and the spikes start to smooth out, leaving a long tail on the high-frequency side. This trend indicates that the hidden states in deeper layers are more locally correlated, which leaves space for Fourier transform to squeeze out the redundancies.

\section{Fourier Transformer}
\subsection{Model Architecture}
The overall architecture of the Fourier Transformer is depicted in Figure \ref{fig: model_arch}. In general, we insert \textit{spectral filters} between layers in Transformer, inside which we use DCT and IDCT to downsample sequence lengths. Multiple spectral filters can work together to split Transformer layers into different blocks, thus progressively reduce sequence lengths. We leave the self-attention intact in order to retain its ability to inherit pretrained weights.

As for the spectral filter, it consists of three steps, i.e., \textit{transform}, \textit{truncate}, and \textit{reverse}. Formally, for an incoming hidden sequence $\{\boldsymbol{h}_{n}\},0 < n < N-1$ that contains $N$ hidden vectors $\boldsymbol{h}_n \in \mathbb{R}^{D}$ where $D$ is the hidden size of the model, the spectral filter first transforms it into frequency domain through 1D-DCT:
\begin{equation}
     \{\boldsymbol{y}_{k}\} = DCT(\{\boldsymbol{h}_{n}\}), \qquad 0 < k < N-1
\end{equation}
Note that the DCT is independently applied on all dimension in $\{\boldsymbol{h}_n\}$, therefore only transforming along the time dimension. 

Next, $\{\boldsymbol{y}_{k}\}$ is truncated by chopping off the trailing dimensions on the high frequency side. For sequences of different lengths, we fix a ratio $r \in (0, 1)$, which is a hyperparameter, to determine the number of frequency components to retain. Thus the length of $\{\boldsymbol{y}_{k}\}$ is truncated from $N$ into $\left\lceil rN \right\rceil$. \footnote{we've played with various ways of truncation here, such as cutting off the low-frequency components, cutting off the ones with lower mean amplitudes, removing the DC components and re-normalize the sequence, subtracting a common value on all components and re-normalize, or retaining the components corresponding to the spikes. Interestingly, the rather classical way of simply chopping off high frequency ones turns out to work the best.}

Finally, the resulting shorter sequence $\{\boldsymbol{y}_{k}\}, 0<k<\left\lceil rN \right\rceil -1$ can be transformed back to time domain through IDCT, yielding a shorter sequence of $\{\boldsymbol{\tilde{h}}_{n}\}$:
\begin{equation}
     \{\boldsymbol{\tilde{h}}_{n}\} = IDCT(\{\boldsymbol{y}_{k}\}), \qquad 0 < n < \left\lceil rN \right\rceil - 1
\end{equation}
Again, IDCT is also conducted in the time dimension only. The resulting shorter hidden states are passed towards upper layers. 

Depending on the type of tasks, the subsequent parts differs. We'll elaborate them in encoder-only and encoder-decoder settings.

\paragraph{Encoder-Only Setting} For encoder-only tasks such as text classification, the final output of the encoder is expected to be a fixed-size vector, which is then fed into logistic regression for class probability predictions. In this work, while the model is trained from scratch, we simply use a mean pooling over the whole output sequence to yield this vector; otherwise when the model inherits a \texttt{[CLS]} token from pretrained models, we use the embedding at that token instead.

\paragraph{Encoder-Decoder Setting} For language generation tasks that involve both an encoder and a decoder, there is an encoder-decoder attention that attends to the encoder states at each decoder step. However, the encoder-decoder attention requires fine-grained positional resolution in order to work well. As a result we follow \citet{dai2020funnel} to upsample the shorter sequences back to their original length, and add the upsampled hidden sequences at all blocks together before feeding them to the decoder. More specifically, we use the parameter-free nearest neighbor interpolation for upsampling, and we re-normalize the sequence after adding the upsampled sequences.

\subsection{Further Pretraining}
\label{sec: further_pretrain}
Since the DCT is reversible through IDCT, the proposed model seamlessly approximates the vanilla Transformer as $r$ goes up. Figure \ref{fig: ablation} shows that while fine-tuning directly on BART \cite{lewis2019bart} weights, the model performs comparatively well when up to $70\%$ frequency components are truncated. Nevertheless, since the upsampling and addition of upsampled sequences still differs from the original Transformer, we can still squeeze the last drop out by applying a tiny amount of further pretraining before fine-tuning, and further improve the model performance. 
This type of further pretraining is much more favourable than a customized pretraining from scratch, which could take massive amount of computation resources. 

As a concrete example, further pretraining our model on BART-Large consumes around 10GB of data and takes around 4 days on 2 NVidia A100 GPUs, while pretraining BART from scratch needs to consume 160GB data, taking roughly 1000 days with the same devices. 
Compared to a customized pre-training from scratch, leveraging BART weights and further pretraining takes 2 magnitudes less computation resources, while still able to bring the model to similar or even better performance.

\subsection{Complexity Analysis}
\label{sec: complexity}
For a standard Transformer layer with model dimension $D$, which consists of self-attention and 2 feed-forward layers, the time and memory complexity of processing an input sequence with length $N$ is $O(N^{2}D + ND^{2})$ and $O(N^2 + ND)$, respectively. With FFT operator 
our model could compress the sequence length from $N$ to $\left\lceil rN \right\rceil$ within $O(N\log N)$ time complexity. Hence the Fourier Transformer enjoys time and memory complexity of $O(r^2N^2D + rND^{2} + N\log N)$ and $O(r^2N^2 + rND)$ every time the sequence length is reduced. Actually, given the parallel implementation of FFT, the additional $O(N\log N)$ time complexity term is negligible compared to the other two terms. The speedup could get even more impressive when the sequence length is relatively long. We refer the readers to Section \ref{sec: lra} for more details. 

\section{Experiments}
\label{sec: exp}
In this section, we experiment with our model in both of the two encoder-only and encoder-decoder settings in various datasets that involves long sequences.

\begin{table*}[ht]
\centering
\begin{tabular}{@{}lccccc | c@{}}
\toprule
\textbf{Models} & \textbf{ListOps} & \textbf{Text} & \textbf{Retrieval} & \textbf{Image} & \textbf{Pathfinder} & \textbf{Avg.} \\ \midrule
Transformer \citep{vaswani2017attention}           & 36.37 & 64.27 & 57.46 & 42.44 & 71.40 & 54.39 \\ 
Longformer \citep{beltagy2020longformer}            & 35.63 & 62.85 & 56.89 & 42.22 & 69.71 & 53.46 \\ 
Linformer \citep{wang2020linformer}             & 35.70 & 53.94 & 52.27 & 38.56 & 76.34 & 51.36 \\ 
Reformer \citep{kitaev2020reformer}              & 37.27 & 56.10 & 53.40 & 38.07 & 68.50 & 50.67 \\ 
Synthesizer \citep{tay2021synthesizer}           & 36.99 & 61.68 & 54.67 & 41.61 & 69.45 & 52.88 \\ 
BigBird \citep{zaheer2020big}               & 36.05 & 64.02 & 59.29 & 40.83 & 74.87 & 55.01 \\ 
Performer \citep{choromanski2020masked}             & 18.01 & 65.40 & 53.82 & 42.77 & 77.50 & 51.41 \\ 
FNet \citep{lee2021fnet}                  & 35.55 & 65.11 & 59.61 & 38.67 & 77.80 & 55.30 \\ 
Nystr\"{o}m \citep{xiong2021nystromformer}               & 37.15 & 65.52 & 79.56 & 41.58 & 70.94 & 58.95 \\ 
Luna-256 \citep{ma2021luna}              & 37.25 & 64.57 & 79.29 & 47.38 & 77.32 & 61.24 \\ 
FSAT \citep{zhuang2022long}                  & \textbf{46.85} & 65.95 & 81.11 & 49.97 & 77.32 & 64.24 \\ \midrule
Fourier Transformer (ours) &   40.73    &   \textbf{75.02}    &   \textbf{85.35}    &   \textbf{53.17}    &    \textbf{83.43}   &  \textbf{67.54}     \\ \bottomrule
\end{tabular}
\caption{The results on LRA benchmark. We report classification accuracy for each task and average accuracy across all tasks. Results from Longformer to Performer are from \citet{tay2020long}, the rest are fetched from their respective papers. For FSAT model on Text task, we only consider the result without convolutions.}
\label{tab: lra}
\end{table*}
\begin{table*}[ht]
\centering
\begin{tabular}{@{}l|cccc|cccc@{}}
\toprule
                      & \multicolumn{4}{c|}{Steps per second $\uparrow$ }    & \multicolumn{4}{c}{Peak Memory Usage $\downarrow$}      \\ \midrule
Model                 & 1K            & 2K               & 3K               & 4K                & 1K               & 2K               & 3K               & 4K               \\ \midrule
Transformer           & 1.0x          & 1.0x             & 1.0x             & 1.0x              & 1.0x             & 1.0x             & 1.0x             & 1.0x             \\ 
Reformer              & 0.5x          & 0.4x             & 0.7x             & 0.8x              & 0.56x            & 0.37x            & 0.28x            & 0.24x            \\ 
BigBird               & 0.9x          & 0.8x             & 1.2x             & 1.1x              & 0.91x            & 0.56x            & 0.4x             & 0.3x             \\ 
Synthesizer           & 1.1x          & 1.2x             & 2.9x             & 1.4x              & 0.76x            & 0.75x            & 0.74x            & 0.74x            \\ 
FSAT                  & 1.1x          & 1.5x             & 2x               & 2.5x              & 0.53x            & 0.27x            & 0.21x            & 0.16x            \\ 
Linformer             & 1.2x          & 1.9x             & 3.7x             & 5.5x              & 0.44x            & 0.21x            & 0.18x            & \textbf{0.1}x             \\ 
Performer             & 1.2x          & 1.9x             & 3.8x             & 5.7x              & 0.44x            & 0.22x            & \textbf{0.15}x            & 0.11x            \\ \midrule
Fourier Transformer (ours) & \textbf{6.9}x & \textbf{12.2}x & \textbf{16.8}x & \textbf{17.7}x & \textbf{0.23}x & \textbf{0.19}x & 0.18x & 0.18x \\ \bottomrule
\end{tabular}
\caption{The speed and memory consumption on LRA benchmark over \textbf{Text} task with input lengths of 1K, 2K, 3K and 4K. The results from Reformer to Performer are from \citet{zhuang2022long}. The speed and memory consumption are listed as the rate w.r.t. the vanilla Transformer.}
\label{tab: lra_time}
\end{table*}

\subsection{Encoder-only Tasks}  \label{sec: lra}
To test our model's ability on encoder-only tasks, we choose the 5 tasks in the widely-used Long Range Arena (LRA) benchmark \citep{tay2020long}. LRA is designed for evaluating efficient transformers under long-context scenario, with the input sequence lengths ranging from 1K to 8K. The datasets in LRA come from rich sources, including natural languages, image pixels, math expressions etc. More specifically, they are:
\paragraph{ListOps} A dataset of math expressions that asks the model to calculate the output value of a math expression with sequence lengths up to 2K.
\paragraph{Text} A byte-level text classification task, with a fixed sequence length 4K which requires the model to deal with compositionality.
\paragraph{Retrieval} A byte-level document retrieval task with a maximum length of 8K which test the model's ability to compress long sequences.
\paragraph{Image} An image classification task of which requires the model to learn the 2D spatial relations between input pixels by sequentially reading the pixels. The sequence length is fixed to 1K.
\paragraph{Pathfinder} An synthetic image classification task with a fixed input length of 1K which requires the model to capture long-range spatial dependencies.

\subsubsection{Implementation Details}
\label{sec: enc imp details}
We run experiments on the LRA benchmark closely following the configurations in \citep{tay2020long}, including data pre-processing, data split, model architecture, hyperparameters (number of layers, hidden dimensions, etc.). We evaluate in terms of classification accuracy. Our implementation is based on \citep{xiong2021nystromformer}. For the sake of simplicity, we report the results of our model over the five tasks with the same compression budget. We aggressively reduce $80\%$ of the input sequence length at the first layer. 

\subsubsection{Performance \& Efficiency}
The results on the aforementioned 5 tasks are summarized in Table \ref{tab: lra}. We compare Fourier Transformer with a bunch of previously published Transformer-based models, and it achieves new state-of-the-art results on four out of the five tasks. Our proposed model improves over the previous SOTA model \citep{zhuang2022long} on \textit{Text}, \textit{Retrieval}, \textit{Image} and \textit{Pathfinder} by 9.07\%, 4.24\%, 3.20\%, 6.11\% absolute value respectively, which is a big margin. Notably, our model doesn't beat FSAT \citep{zhuang2022long} on the ListOps task and ranks the 2nd in the list. We conjecture that it's because math expression values are more sensitive to individual tokens in the sequence, thus is more sensitive to downsampling. 

Next, taking the byte-level text classification task (the Text dataset) as a testbed, we quantitatively evaluate the time and memory efficiency of our model and the other competing models on various input lengths. The results are summarized in Table \ref{tab: lra_time}. Note that, due to the limitation of GPU memory for the vanilla Transformer, results on 1K, 2K and 3K lengths are run with a batch size of 32, and 4K are with a batch size of 16. We calculate the corresponding rates of our model w.r.t. vanilla Transformer on identical batch settings, and timed on an NVidia A100-80G GPU. Compared with other efficient transformers, Fourier Transformer significantly reduces time consumption on both short and long sequences, leaving the other model behind by a large margin, while keeping a steady memory savings as the sequence length grows.

\subsection{Encoder-Decoder Tasks}
The model for encoder-decoder tasks are equipped with a decoder to perform text generation. For this setting, we choose two long-text datasets in summarization and question answering tasks, i.e., CNN/DailyMail \citep{hermann2015teaching} and 
ELI5 \citep{eli5_lfqa}, with average sequence lengths at 0.8K and 5K, respectively.


\paragraph{CNN/DailyMail} A summarization dataset containing over 280K news articles (766 token counts on average) from news stories in CNN and Daily Mail websites paired with human-generated summaries (53 token counts on average). We follow the conversion and evaluate the performance in terms of Rouge scores (Rouge-1, Rouge-2, Rouge-L) \citep{lin2004rouge}.
\paragraph{ELI5} A question answering dataset containing over 270K complex, diverse and paragraph-length question-answer pairs gathered from subreddits, the average number of tokens for input and target are 5140 and 693 respectively. Following the conversion, we evaluate it in both Rouge-L and F1 scores.

\subsubsection{Implementation Details}
\label{sec: enc-dec imp details}
Since on both the two datasets pretrained models leave a large gap over non-pretrained ones, it makes less sense to report results without pretraining. Thus, we report results of our model inheriting BART-large \citep{lewis2019bart} weights. We generally test two settings, which is: 1) directly fine-tune our model on the dataset, and 2) conduct further pretraining before fine-tuning. For convenience, we call them \emph{Fourier-BART} and \emph{Fourier-BART-FP} respectively in the rest of the paper. 

Fourier-BART has the same architecture as BART-large. It simply adopts a 2-block design, the first block contains the first 2 consecutive transformer layers, the rest 10 layers belong to the second block. 
For CNN/DailyMail, $50\%$ of the frequency components are truncated, while for ELI5 $70\%$ are truncated since it has much longer sequence lengths. 

Fourier-BART-FP has the same setting as Fourier-BART, except that before fine-tuning on downstream tasks it is further pretrained for 1 epoch on 10GB of text with the original BART pretraining objectives. The text is randomly sliced from the Pile \citep{gao2020pile} corpus.

\subsubsection{Performance \& Efficiency}
\paragraph{CNN/DailyMail} On summarization task, inside the scope of efficient models, we compare our model with BigBird \citep{zaheer2020big}, ST-MoE \citep{zoph2022designing} and Switch Transformer \citep{fedus2021switch}, which are strong baselines from recent literature. Both ST-MoE and Switch Transformer targeted at activating only part of the parameters to improve the efficiency. Bigbird approximates full attention matrix with a sparse one to improve on FLOPs. In addition, we put the standard BART \citep{lewis2019bart} performance as baseline. 

The results are listed in Table \ref{tab: cnn}. Our proposed Fourier-BART successively leverages the advantage of BART, achieving a performance at the level of pretrained model. With the tiny amount of further pretraining, it achieves the best performance among all competitors. Note that Fourier-BART is built upon BART and sharing the same model size with BART-400M with much less computation, however it is able to outperform the standard BART-400M with a sensible margin.

\begin{table}[ht]
\centering
\begin{tabular}{@{}lccc@{}}
\toprule
Model                           & R-1 & R-2 & R-L \\ \midrule
BART-400M                       & 44.16   & 21.28   & 40.90   \\ 
ST-MOE-L-770M                      & -       & 20.7    & -       \\ 
Switch Trans.-223M              & -       & 19.6    & -       \\ 
BigBird-Large                   & 43.84   & 21.11   & 40.74   \\ 
\midrule
Fourier-BART-400M      &  44.65   &  21.48       &   41.30      \\ 
Fourier-BART-FP-400M     &  \textbf{44.76}   &  \textbf{21.55}       &   \textbf{41.34}      \\\bottomrule
\end{tabular}
\caption{Rouge scores on CNN/DailyMail. The results are all fetched from their respective papers. The R-1 and R-L of ST-MOE and Switch Transformer are not reported in their paper. The number after model name denotes the model size. The model size for BigBird is not mentioned in their paper unfortunately.}
\label{tab: cnn}
\end{table}

\begin{table}[ht]
\centering
\begin{tabular}{@{}lcccc@{}}
\toprule
Model                                    & RL & F1 \\ \midrule
LayerDrop-240M      & 23.4   & -   \\
E-MCA-240M               & 24.0   & -   \\
c-REALM*-596M      & 23.2    & 22.9   \\
EMAT*-446M            & 20.91   & 19.03   \\
KID*-406M            & 26.3   & -   \\
BART-large-400M                               & 26.8   & 26.6   \\
\midrule
Fourier-BART-400M                                    & 26.2   & 25.98   \\
Fourier-BART-FP-400M                          & \textbf{26.9}   & \textbf{26.73}   \\ \bottomrule
\end{tabular}
\caption{Model performance on ELI5. The results from E-MCA to KID are fetched from their respective papers. * denotes results using the Kilt benchmark \citep{petroni2020kilt}, which has smaller dev and test sets.}
\label{tab: eli5}
\end{table}

As for efficiency, it is almost impossible to reproduce all the models listed in Table \ref{tab: cnn} and investigate their efficiency, so we choose to only evaluate the standard BART-400M and proposed Fourier-BART-400M in terms of FLOPs. 
As elaborated in Section \ref{sec: enc-dec imp details}, we remove 50\% from the hidden sequence on the third transformer layer, although the two models have the exact same size, the FLOPs invested in the standard BART-400M is 1.6 times of Fourier-BART-400M. 
Due to the upsampling and the auto-regressive decoding, the overall reduction in computation is not as significant as those on LRA.

\paragraph{ELI5}
On question answering task, we compare our model with the LayerDrop \citep{fan2019reducing}, E-MCA \citep{fan2019using}, c-REALMS \citep{krishna2021hurdles}, EMAT \citep{wu2022efficient} and KID \citep{liu2022knowledge}. To provide a fair comparison, the result of BART-large is our reproduced one on the bleeding-edge version of fairseq \cite{ott2019fairseq}, which is much higher than the results reported in the original BART paper. Note that here we are even comparing with performance-sensitive models, as in the list only EMAT and LayerDrop are focusing on reducing complexity. As shown in Table \ref{tab: eli5}, our Fourier-BART-FP has surpassed all the competing models on both Rouge-L and F1 scores. 

As for efficiency, when removing 70\% of the frequency components (elaborated in Section \ref{sec: enc-dec imp details}), the FLOPs invested in the standard BART is 1.9 times of Fourier-BART.

\begin{figure}[ht]
\centering
\includegraphics[width=\linewidth]{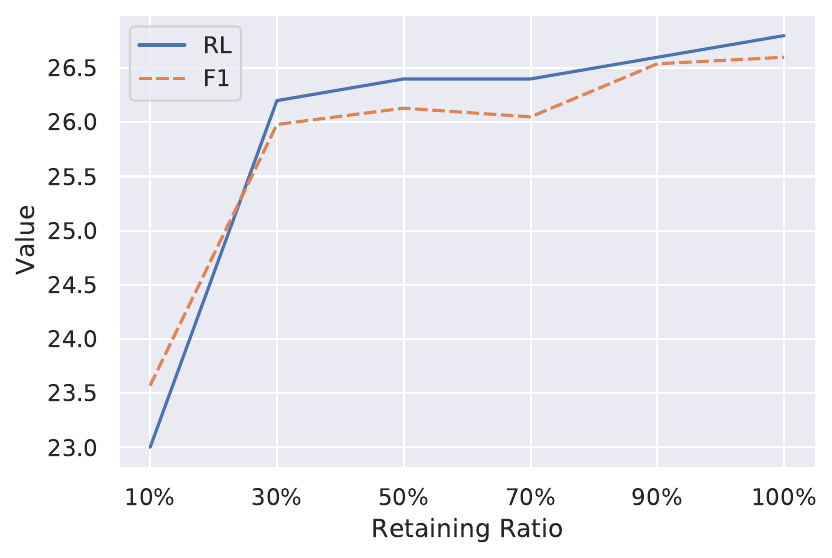} 
\caption{R1, R2, RL and F1 on ELI5. x-axis stands for the retraning ratio $r$.}
\label{fig: ablation}
\end{figure}

\subsection{Analysis on Retaining Ratio $r$}
An important question that arises is how sensitive the model is w.r.t. the ratio of retaining frequency components. To investigate this, we experiment our model in ELI5 dataset. by sweeping $r$ from $0.1$ to $1$. We didn't conduct further pretraining on each setting due to computation limit. Results are shown in Fig \ref{fig: ablation}. The performance remains pretty good up until less than $30\%$ of frequency components are retained. When we try to truncate more components passing that ratio, the performance starts to drop significantly. This is a fairly satisfying result that shows the model performs reliably stable in a wide range of reasonable $r$'s.


\section{Conclusion}
In this work, we introduce the discrete cosine transformation to progressively downsample the hidden states in the Transformer model by leveraging the local correlations between hidden states in upper layers. Our approach is able to significantly reduce the computation required by the vanilla Transformer, while being able to achieve even better performance in various tasks. Moreover, it is able to inherit the pretrained model weights, which is an notable advantage over most efficient Transformers.

\section{Limitations}
Although our approach exhibits great speedups in encoder-only settings, it doesn't yield as impressive speedups in encoder-decoder setting. This is due to the autoregresive decoding steps in the decoder, that has to be conducted sequentially. Accelerating that with DCT requires to incrementally update DCT outputs step by step based on outputs of previous timesteps, which is theoretically possible but not easy to optimize its efficiency. We plan to further accelerate it in this direction in future work.

\section*{Acknowledgement}
This work was sponsored by the National Natural Science Foundation of China (NSFC) grant (No.
62106143), and Shanghai Pujiang Program (No. 21PJ1405700). 

\bibliography{custom}
\bibliographystyle{acl_natbib}




\end{document}